\documentclass[conference]{IEEEtran}
\IEEEoverridecommandlockouts

\usepackage[T1]{fontenc}
\usepackage{amsmath,amssymb,amsfonts}
\usepackage{graphicx}
\usepackage{booktabs}
\usepackage{array}
\usepackage{multirow}
\usepackage{makecell}
\usepackage{xcolor}
\usepackage{algorithm}
\usepackage{algpseudocode}
\usepackage{tikz}
\usetikzlibrary{arrows.meta, positioning, fit, calc, backgrounds}
\usepackage{pgfplots}
\pgfplotsset{compat=1.18}
\usepgfplotslibrary{groupplots}
\usepackage{url}
\usepackage{balance}
\usepackage{microtype}
\usepackage{hyperref}
\usepackage{cite}

\definecolor{coral}{HTML}{E74C3C}
\definecolor{forest}{HTML}{27AE60}
\definecolor{ocean}{HTML}{2980B9}
\definecolor{amber}{HTML}{F39C12}
\definecolor{slate}{HTML}{34495E}
\definecolor{violet}{HTML}{8E44AD}
\definecolor{teal}{HTML}{16A085}

\hypersetup{colorlinks=true, linkcolor=ocean, citecolor=ocean, urlcolor=ocean}

\newcommand{\semb}{s_{\mathrm{emb}}}
\newcommand{\sstr}{s_{\mathrm{str}}}
\newcommand{\sent}{s_{\mathrm{ent}}}
\newcommand{\conv}{\circledast}
\DeclareMathOperator{\cossim}{cos}

\begin{document}

\title{Verification Without Sufficiency: Per-Chunk Filtering Fails\\
on Multi-Hop RAG, and Decomposition Repairs It}

\author{\IEEEauthorblockN{Randhir Kumar}
\IEEEauthorblockA{Independent Researcher\\
\texttt{randhir2709vns@gmail.com}}}

\maketitle

% ============================================================================
\begin{abstract}
Verification for retrieval-augmented generation usually scores each retrieved
chunk and drops the ones that fail. We show this cannot work for multi-hop
questions, and show what does. Per-chunk scoring assumes one chunk is a
sufficient premise for the answer. Multi-hop questions are built so that none
is, and the paragraph carrying the answer is the one the question does not name.
Entailment scoring reaches 0.643, 0.523 and 0.560 AUC on HotpotQA,
2WikiMultihopQA and MuSiQue, against 0.951 on single-hop SQuAD. Seven controls
rule out model capacity, premise length, hypothesis template, decision
threshold, retriever, answer-matching criterion and prompt. End to end across
three datasets, three generator sizes and two prompts, per-chunk gating is
significantly worse than not filtering at all in every cell, and its penalty
grows with generator capability. The repair is to condition verification on the
decomposed sub-question rather than the original query. Using MuSiQue's gold
decomposition, entailment on a later hop rises from 0.546, which is chance, to
0.840, a paired lift of $+0.355$ with a bootstrap interval of $[0.331, 0.382]$.
An off-the-shelf Qwen2.5-7B decomposer, given the question and the top retrieved
paragraph, reaches 0.637 and captures 31\% of that ceiling; decomposing without
retrieval reaches 0.533, below the original question. Iterative retrieval
systems already produce such decompositions and discard them before verifying.
\end{abstract}

\begin{IEEEkeywords}
Retrieval-augmented generation, natural language inference, multi-hop question
answering, evidence sufficiency, negative results.
\end{IEEEkeywords}

% ============================================================================
\section{Introduction}
\label{sec:intro}

Retrieval-augmented generation grounds a language model in a corpus by
retrieving passages and conditioning generation on them~\cite{lewis2020retrieval,
gao2023survey}. Retrievers make mistakes, and a generator handed a topically
adjacent but factually unrelated passage will use it anyway. The standard
response is a verification step: score each retrieved chunk, keep the good ones,
drop the rest. CRAG~\cite{yan2024corrective} trains an evaluator that labels
chunks correct, incorrect or ambiguous. Self-RAG~\cite{asai2023self} trains the
generator to emit reflection tokens. Faithfulness metrics such as
RAGAS~\cite{es2024ragas} score entailment between answer and context.

We built such a verifier with three signals, embedding cosine similarity,
natural language inference entailment, and a structural role-filler score based
on Holographic Reduced Representations~\cite{plate1995holographic}. Before
evaluating end to end we measured the thing a verifier has to do: separate gold
evidence from distractors. On HotpotQA the answer was 0.887 for embedding
similarity, 0.643 for entailment and 0.620 for the structural score. Since
embedding similarity is what the retriever already computes, the two signals we
added contributed nothing.

This paper explains why, and the explanation turns out not to be about our
implementation. Per-chunk verification treats each chunk as a sufficient premise
for the answer. Multi-hop questions are constructed so that no single retrieved
paragraph is sufficient, and the paragraph that carries the answer is the one
the question does not name. A verifier conditioned on the question is therefore
strongest on the evidence already in hand and weakest on the evidence being
sought.

\vspace{2pt}
We establish this with seven measurements, each closing an alternative account:

\begin{enumerate}\itemsep2pt
  \item \textbf{It is not one dataset.} Entailment reaches 0.643, 0.523 and
        0.560 AUC on HotpotQA, 2WikiMultihopQA and MuSiQue
        (Section~\ref{sec:weak}).
  \item \textbf{The failure has a direction.} All three signals prefer the
        paragraph named in the question. On comparison questions, where both
        entities are named, the deficit disappears
        (Section~\ref{sec:direction}).
  \item \textbf{It deepens with hop count.} On MuSiQue, AUC falls monotonically
        from 2-hop to 4-hop questions (Section~\ref{sec:hops}).
  \item \textbf{It is sufficiency, not length.} Both gold paragraphs together
        reach 0.881 AUC; a gold paired with a distractor scores 0.127 against
        0.540, at greater length (Section~\ref{sec:sufficiency}).
  \item \textbf{It is not the task.} On single-hop SQuAD the same pipeline
        reaches 0.951 (Section~\ref{sec:singlehop}).
  \item \textbf{No threshold rescues it.} At the most permissive threshold
        tested, 83\% of gold paragraphs are already rejected
        (Section~\ref{sec:threshold}).
  \item \textbf{It is not the embedder.} The deficit holds for three embedding
        models (Section~\ref{sec:threshold}).
\end{enumerate}

Section~\ref{sec:endtoend} measures what this costs. Across three datasets (500, 500 and 259 evaluated questions),
three generator sizes and two prompts, per-chunk gating is the worst of seven
selectors in every cell, and the penalty grows with generator capability.

Section~\ref{sec:decomp} shows what repairs it. Conditioning the hypothesis on
the decomposed sub-question rather than the original query lifts entailment on a
later hop from 0.546 to 0.840, and hop count stops mattering.
This is the positive result of the paper, and it says where verification for
multi-hop retrieval should be built: on the same decomposition that iterative
retrieval already produces.

% ============================================================================
\section{Related Work}
\label{sec:related}

\paragraph{Corrective and self-critiquing RAG}
CRAG~\cite{yan2024corrective} adds a trained retrieval evaluator that branches
generation on a per-chunk verdict. Self-RAG~\cite{asai2023self} trains the model
to critique its own retrieval during decoding. Both operate on individual
passages and are evaluated on tasks that are largely single-hop. Our results
suggest the per-chunk granularity is the part that does not transfer.

\paragraph{NLI for grounding}
Entailment models~\cite{bowman2015snli, williams2018mnli, he2023debertav3} are
widely used to check whether generated text is supported by retrieved
context~\cite{es2024ragas}. That use is post-hoc and set-level: the premise is
the whole context the generator saw. Moving the check earlier, to filter chunks,
changes the premise to a single passage. It is that change our measurements
isolate.

\paragraph{Multi-hop retrieval}
HotpotQA~\cite{yang2018hotpotqa}, 2WikiMultihopQA~\cite{ho20202wiki} and
MuSiQue~\cite{trivedi2022musique} supply gold supporting paragraphs alongside
hard distractors, and MuSiQue is constructed specifically to resist single-hop
shortcuts. Iterative methods such as Self-Ask~\cite{press2023selfask} and
IRCoT~\cite{trivedi2023ircot} rewrite the query between hops because the second
hop is not reachable from the original question. Our findings say the same thing
about verification that those methods say about retrieval, and we quantify it.

\paragraph{Holographic representations}
HRRs encode role-filler structure in fixed-width vectors via circular
convolution~\cite{plate1995holographic, plate2003book}, a construction shared
with the wider vector-symbolic literature~\cite{kanerva2009hyperdimensional,
nickel2016hole}. We are not aware of prior work using HRR as a chunk
verification signal, and Section~\ref{sec:hrr} reports why it does not work.

% ============================================================================
\section{Setup}
\label{sec:setup}

\subsection{Three verification schemes}

\begin{figure*}[t]
\centering
\resizebox{\textwidth}{!}{%
\begin{tikzpicture}[
  font=\small,
  chunk/.style={rectangle, rounded corners=2pt, draw=slate, thick,
                minimum width=1.5cm, minimum height=0.55cm, align=center,
                font=\footnotesize},
  gold/.style={chunk, fill=forest!22},
  dist/.style={chunk, fill=slate!8},
  scorer/.style={rectangle, rounded corners=3pt, draw=ocean, very thick,
                 fill=ocean!12, minimum width=1.9cm, minimum height=0.8cm,
                 align=center, font=\footnotesize},
  arr/.style={-{Stealth[length=2mm]}, thick, draw=slate},
  ttl/.style={font=\small\bfseries, text=slate},
  note/.style={font=\scriptsize, text=slate!85, align=center},
]
% ---------------- panel A: per chunk
\begin{scope}[shift={(0,0)}]
  \node[ttl] at (2.1,2.5) {(a) per chunk};
  \node[note] at (2.1,2.05) {premise = one paragraph};
  \node[gold] (a1) at (0,1.2) {$c_1$ hop 1};
  \node[gold] (a2) at (0,0.5) {$c_2$ hop 2};
  \node[dist] (a3) at (0,-0.2) {$c_3$};
  \node[dist] (a4) at (0,-0.9) {$c_4$};
  \foreach \i in {1,2,3,4} {\draw[arr] (a\i.east) -- ++(0.55,0);}
  \node[scorer] (as) at (2.6,0.15) {$\sent(q,c_i)$};
  \node[note, text=coral] at (2.6,-0.95) {$c_2$ alone does not\\entail the answer};
  \draw[arr] (as.east) -- ++(0.6,0) node[right, font=\footnotesize] {keep / drop};
\end{scope}
% ---------------- panel B: set level
\begin{scope}[shift={(7.6,0)}]
  \node[ttl] at (2.1,2.5) {(b) set level};
  \node[note] at (2.1,2.05) {premise = every pair, $\binom{k}{2}$ calls};
  \node[gold] (b1) at (0,1.2) {$c_1$};
  \node[gold] (b2) at (0,0.5) {$c_2$};
  \node[dist] (b3) at (0,-0.2) {$c_3$};
  \node[dist] (b4) at (0,-0.9) {$c_4$};
  \begin{scope}[on background layer]
    \node[draw=amber, thick, dashed, rounded corners=3pt, fill=amber!8,
          fit=(b1)(b4), inner sep=4pt] {};
  \end{scope}
  \draw[arr] (0.85,0.15) -- ++(0.6,0);
  \node[scorer] (bs) at (2.6,0.15) {$\sent(q,c_i\!\oplus\!c_j)$};
  \node[note, text=amber!80!black] at (2.6,-0.95) {argmax over 10 pairs\\lands correctly 44\% of the time};
  \draw[arr] (bs.east) -- ++(0.6,0) node[right, font=\footnotesize] {best pair};
\end{scope}
% ---------------- panel C: conditional
\begin{scope}[shift={(15.6,0)}]
  \node[ttl] at (2.3,2.5) {(c) conditional (ours)};
  \node[note] at (2.3,2.05) {anchor on hop 1, $k-1$ calls};
  \node[gold, fill=forest!45] (c1) at (0,1.2) {$c_1$ anchor};
  \node[gold] (c2) at (0,0.5) {$c_2$};
  \node[dist] (c3) at (0,-0.2) {$c_3$};
  \node[dist] (c4) at (0,-0.9) {$c_4$};
  \node[note, text=forest!70!black, left=1pt of c1] {$\arg\max\semb$\\92\% gold};
  \foreach \i in {2,3,4} {\draw[arr] (c\i.east) -- ++(0.55,0);}
  \draw[arr, draw=forest, very thick] (c1.east) -- ++(0.55,0) |- (2.6,0.9);
  \node[scorer, draw=forest, fill=forest!12] (cs) at (2.8,0.05)
        {$\sent(q,c_1\!\oplus\!c_j)$};
  \node[note, text=forest!70!black] at (2.8,-0.95) {recall $0.755$ vs $0.683$\\at 4 calls not 10};
  \draw[arr] (cs.east) -- ++(0.6,0) node[right, font=\footnotesize] {completion};
\end{scope}
\end{tikzpicture}}
\caption{Three ways to score retrieved evidence. Only the premise changes.
(a) Per-chunk scoring asks whether one paragraph entails the answer, which for
the second hop of a multi-hop question it cannot. (b) Set-level scoring restores
sufficiency but searches blindly. (c) Conditional scoring anchors on the
paragraph the embedding signal identifies reliably, then asks which paragraph
completes it. Numbers are measured on HotpotQA, 500 questions.}
\label{fig:schemes}
\end{figure*}
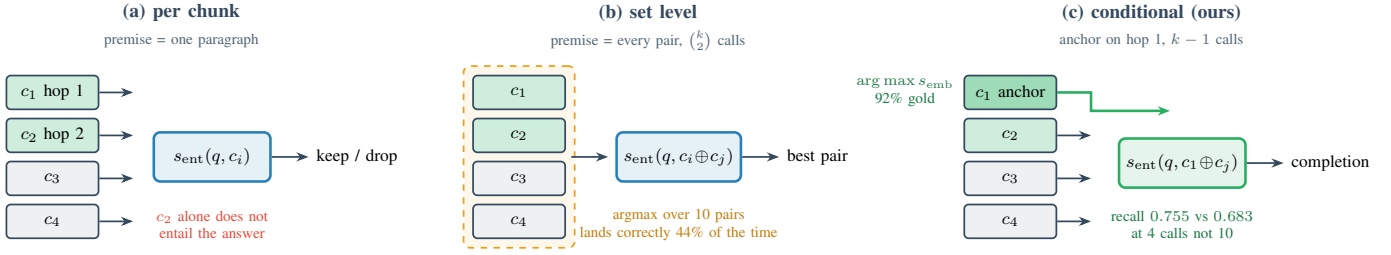

Figure~\ref{fig:schemes} contrasts the three ways of scoring evidence that this
paper compares. All of them use the same models; they differ only in what
constitutes a premise.

\subsection{The three signals}

Given a question $q$ and a chunk $c$:

\paragraph{Embedding} Cosine similarity between bi-encoder embeddings
(BAAI/bge-small-en-v1.5, 33\,M), rescaled to $[0,1]$:
$\semb(q,c)=\tfrac12\big(1+\cossim(E(q),E(c))\big)$.

\paragraph{Entailment} A rule-based operator $\mathcal{H}$ converts the
interrogative into a declarative proposition and a cross-encoder scores
entailment with the chunk as premise,
$\sent(q,c)=P(\text{entail}\mid c,\mathcal{H}(q))$. Getting $\mathcal{H}$ right
matters more than it appears. A hypothesis phrased as a statement about the
passage, of the form ``this text contains information about $q$'', is out of
distribution for a model trained on SNLI or MNLI, where premise and hypothesis
are both object-level claims, and it collapses into a topicality detector. We
generate object-level propositions with the answer slot underspecified: ``Who
directed Inception?'' becomes ``a person directed Inception.'' We report
\textsc{slot} as above and \textsc{oracle}, which fills the slot with the gold
answer and therefore upper-bounds any deployable version of the signal.

\paragraph{Structure} Role-filler pairs over
$\{\textsc{subj},\textsc{obj},\textsc{root},\textsc{ent}\}$ are bound with
circular convolution into a trace and recovered by unbinding:
\begin{align}
T(c) &= \textstyle\sum_{r\in\mathcal{R}_c}\mathbf{v}_r\conv\mathbf{v}_{\phi_c(r)},\\
\sstr(q,c) &= \tfrac{1}{|\mathcal{R}_q|}\textstyle\sum_{r\in\mathcal{R}_q}
\max\!\big(0,\cossim(\mathbf{v}_r^{\dagger}\conv T(c),\mathbf{v}_{\phi_q(r)})\big).
\end{align}
Fillers are whitened before binding, since raw sentence embeddings are
anisotropic and violate the near-orthogonality binding assumes. Whitening
reduces mean pairwise cosine from $0.488$ to $-0.0001$.

\subsection{Data}

Three multi-hop datasets, 500 questions each, sampled with a fixed seed:
HotpotQA~\cite{yang2018hotpotqa} distractor (10 paragraphs per question, 2.0
gold), 2WikiMultihopQA~\cite{ho20202wiki} (10 paragraphs, 2.5 gold) and
MuSiQue~\cite{trivedi2022musique} (20 paragraphs, 2.6 gold, and 2, 3 or 4 hops).
Retrieval is per-question over the supplied candidates, so no global index is
involved. All 10 or 20 paragraphs are scored, giving 20\,000
question-paragraph pairs.

For the single-hop control we use SQuAD v1.1~\cite{rajpurkar2016squad}, with
distractors drawn from other paragraphs of the same Wikipedia article so they
remain hard negatives. 300 questions, nine distractors each.

\subsection{Models and protocol}

Entailment uses \texttt{nli-deberta-v3-base} (184\,M) unless noted; we also test
\texttt{xsmall} (44\,M) and a 435\,M DeBERTa-v3-large. Before use we verified
each model's entailment logit index against two unambiguous NLI pairs; all three
assign the correct label with margins above $0.95$, so no result below is a
label-mapping artefact. Generation uses Qwen2.5-1.5B-Instruct~\cite{yang2024qwen}
with greedy decoding and seed 1337.

We report AUC of separating gold paragraphs from distractors. AUC is
threshold-free, and a verifier that cannot rank gold above distractor cannot be
rescued by tuning a cutoff. Section~\ref{sec:threshold} confirms that directly.

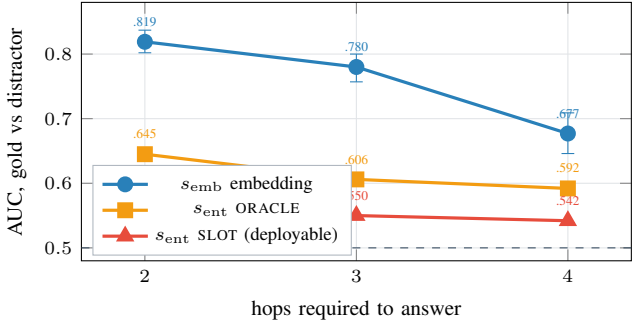
\begin{figure}[t]
\centering
\begin{tikzpicture}
\begin{axis}[
  width=\columnwidth, height=5.0cm,
  xlabel={hops required to answer}, ylabel={AUC, gold vs distractor},
  xmin=1.7, xmax=4.3, ymin=0.48, ymax=0.88, xtick={2,3,4},
  tick label style={font=\scriptsize}, label style={font=\footnotesize},
  grid=both, grid style={slate!15},
  legend style={font=\scriptsize, at={(0.02,0.02)}, anchor=south west,
                draw=slate!40, fill=white},
  nodes near coords, point meta=explicit symbolic,
  every node near coord/.append style={font=\tiny, yshift=2pt},
]
\addplot[very thick, ocean, mark=*, mark size=2.4,
         error bars/.cd, y dir=both, y explicit] coordinates {
  (2,0.819) -= (0,0.017) += (0,0.018) [.819]
  (3,0.780) -= (0,0.023) += (0,0.020) [.780]
  (4,0.677) -= (0,0.031) += (0,0.032) [.677] };
\addlegendentry{$\semb$ embedding}
\addplot[very thick, amber, mark=square*, mark size=2.4] coordinates {
  (2,0.645) [.645] (3,0.606) [.606] (4,0.592) [.592] };
\addlegendentry{$\sent$ \textsc{oracle}}
\addplot[very thick, coral, mark=triangle*, mark size=2.8] coordinates {
  (2,0.590) [.590] (3,0.550) [.550] (4,0.542) [.542] };
\addlegendentry{$\sent$ \textsc{slot} (deployable)}
\addplot[dashed, slate, forget plot] coordinates {(1.7,0.5) (4.3,0.5)};
\node[font=\scriptsize, text=slate, anchor=west] at (axis cs:1.75,0.515) {chance};
\end{axis}
\end{tikzpicture}
\caption{MuSiQue, 500 questions split by hop count (259 / 163 / 78), with
question-stratified bootstrap intervals on the embedding curve. Embedding
similarity degrades significantly with hop count, from 0.819 [0.802, 0.837] at
two hops to 0.677 [0.646, 0.709] at four. The entailment curves also fall but
their intervals overlap, so we report the trend without claiming it. The
deployable entailment variant is within 0.04 of chance at four hops.}
\label{fig:hops}
\end{figure}

% ============================================================================
\section{The signals are weak on every multi-hop dataset}
\label{sec:weak}

\begin{table}[t]
\centering
\caption{Separating gold paragraphs from distractors. 500 questions per
dataset. \emph{deficit} is AUC on bridge-only gold minus AUC on answer-bearing
gold; positive means the signal prefers the paragraph without the answer.}
\label{tab:auc}
\setlength{\tabcolsep}{4pt}
\begin{tabular}{@{}llccc@{}}
\toprule
\textbf{Dataset} & \textbf{Signal} & \textbf{AUC} &
\makecell{\textbf{answer-}\\\textbf{bearing}} & \makecell{\textbf{bridge}\\\textbf{only}}\\
\midrule
\multirow{3}{*}{HotpotQA}
 & $\semb$                 & \textbf{0.887} & 0.849 & 0.941\\
 & $\sent$ \textsc{oracle} & 0.669 & 0.681 & 0.652\\
 & $\sent$ \textsc{slot}   & 0.643 & 0.635 & 0.655\\
\midrule
\multirow{3}{*}{2Wiki}
 & $\semb$                 & \textbf{0.807} & 0.840 & 0.788\\
 & $\sent$ \textsc{oracle} & 0.614 & 0.656 & 0.589\\
 & $\sent$ \textsc{slot}   & 0.523 & 0.548 & 0.507\\
\midrule
\multirow{3}{*}{MuSiQue}
 & $\semb$                 & \textbf{0.762} & 0.702 & 0.798\\
 & $\sent$ \textsc{oracle} & 0.632 & 0.688 & 0.597\\
 & $\sent$ \textsc{slot}   & 0.560 & 0.590 & 0.541\\
\midrule
\multicolumn{2}{@{}l}{HotpotQA, $\sstr$ chunk whitened} & 0.620 & 0.560 & 0.696\\
\multicolumn{2}{@{}l}{HotpotQA, $\sstr$ sentence whitened} & 0.595 & 0.536 & 0.677\\
\bottomrule
\end{tabular}
\end{table}

Table~\ref{tab:auc} is the result that redirected this work. Embedding
similarity separates gold from distractor at 0.76 to 0.89. Neither signal we
added reaches 0.67 on any dataset, and the best entailment variant is the one
handed the gold answer.

That last point deserves emphasis. \textsc{oracle} is not deployable; it exists
to bound the signal. If entailment cannot exceed 0.669 even when the hypothesis
contains the answer verbatim, no amount of work on hypothesis generation will
help, because hypothesis generation is trying to approximate exactly that.

Scaling the entailment model does not rescue it either. On a 200-pair
calibration subset the 44\,M model reaches 0.590 \textsc{oracle} AUC, the
184\,M model 0.668, and the 435\,M model 0.543. We do not read much into the
ordering at that sample size, but a tenfold parameter increase producing no
improvement is itself informative.

% ============================================================================
\section{The failure has a direction}
\label{sec:direction}

\begin{table}[t]
\centering
\caption{AUC by question type. 2Wiki labels four types, HotpotQA two.
Comparison questions name both entities; compositional and inference questions
hide the second hop behind the first.}
\label{tab:qtype}
\setlength{\tabcolsep}{4pt}
\begin{tabular}{@{}llccc@{}}
\toprule
\textbf{Type ($n$)} & \textbf{Signal} &
\makecell{\textbf{answer-}\\\textbf{bearing}} & \makecell{\textbf{bridge}\\\textbf{only}} & \textbf{deficit}\\
\midrule
\multirow{2}{*}{2Wiki comparison (138)}
 & $\semb$   & 0.974 & 0.976 & $+0.002$\\
 & $\sent$   & 0.695 & 0.493 & $-0.203$\\
\multirow{2}{*}{2Wiki compositional (189)}
 & $\semb$   & 0.708 & 0.979 & $\mathbf{+0.271}$\\
 & $\sent$   & 0.469 & 0.658 & $+0.189$\\
\multirow{2}{*}{2Wiki inference (54)}
 & $\semb$   & 0.683 & 0.972 & $\mathbf{+0.289}$\\
 & $\sent$   & 0.465 & 0.594 & $+0.128$\\
\midrule
\multirow{2}{*}{HotpotQA bridge (543)}
 & $\semb$   & 0.796 & 0.927 & $+0.131$\\
 & $\sent$   & 0.605 & 0.695 & $+0.090$\\
\multirow{2}{*}{HotpotQA comparison (157)}
 & $\semb$   & 0.972 & 0.969 & $-0.003$\\
 & $\sent$   & 0.616 & 0.513 & $-0.103$\\
\bottomrule
\end{tabular}
\end{table}

Multi-hop gold evidence comes in two kinds. In a bridge question such as
\emph{``What government position was held by the woman who portrayed Corliss
Archer in Kiss and Tell?''}, one paragraph is about \emph{Kiss and Tell} and
supplies the bridge entity, and the other is about Shirley Temple and supplies
the answer. The first is named in the question. The second is not, and is
reachable only through the first.

We label a gold paragraph \emph{answer-bearing} if the normalised gold answer
occurs in it, and \emph{bridge-only} otherwise. On HotpotQA, 58\% of gold
paragraphs are answer-bearing and 502 of 700 questions have exactly one.

\paragraph{The entailment result is the surprising one}
The \textsc{oracle} hypothesis contains the gold answer string, and the
answer-bearing paragraph contains that same string. Lexical overlap alone should
push entailment higher there. It does not: 0.681 [0.659, 0.704] on
answer-bearing gold against 0.652 [0.625, 0.679] on bridge-only gold for
HotpotQA, and on 2Wiki and MuSiQue the deployable variant is barely above
chance on either. An entailment model handed the answer cannot reliably tell
which paragraph contains it, because the proposition it is checking requires
both hops.

\paragraph{The embedding result is the expected baseline}
That $\semb$ prefers the bridge paragraph is closer to a restatement of what
embedding similarity measures: the question names the bridge entity, so the
question embedding sits near the paragraph about it. We report it because it
calibrates the entailment numbers, not as a discovery. Its magnitude is
nonetheless large, $+0.271$ on 2Wiki compositional questions, where the bridge
paragraph separates at 0.979 and the answer-bearing paragraph at 0.708.

\paragraph{Comparison questions are the control}
On 2Wiki comparison questions, which name both entities, the embedding deficit
collapses to $+0.002$ and absolute AUC rises to 0.974. HotpotQA reproduces the
contrast, $+0.131$ against $-0.003$. Nothing about the models, the corpus or the
retriever changes between those rows. Only the relationship between the question
and the evidence changes.

\paragraph{A note on aggregation}
The 2Wiki row of Table~\ref{tab:auc} shows a \emph{negative} aggregate deficit,
$-0.052$, while Table~\ref{tab:qtype} shows $+0.271$ and $+0.289$ on its
compositional and inference subsets. There is no inconsistency: 2Wiki contains
138 comparison questions whose deficit is strongly negative, and they pull the
pooled figure below zero. The aggregate is a mixture over question types with
opposite signs and should not be read on its own. HotpotQA, with a smaller
comparison share, does not invert.

\paragraph{A worked example}
\emph{``What government position was held by the woman who portrayed Corliss
Archer in Kiss and Tell?''} The bridge paragraph is titled \emph{Kiss and Tell};
its title appears verbatim in the question. The answer-bearing paragraph is
titled \emph{Shirley Temple}, a name the question never uses. Averaged over the
543 HotpotQA bridge questions, the bridge paragraph is separated from
distractors at 0.927 by embedding and 0.695 by entailment; the answer-bearing
paragraph at 0.796 and 0.605. The gap runs the same way for all three signals.

% ============================================================================
\section{It deepens with hop count}
\label{sec:hops}

MuSiQue labels each question with the number of hops its answer requires, which
turns the mechanism into a dose-response test. If the problem is that
query-conditioned signals cannot see past the first hop, then adding hops should
make it worse.

Figure~\ref{fig:hops} shows it does for the embedding signal, where the
intervals separate cleanly: 0.819 [0.802, 0.837] at two hops against 0.677
[0.646, 0.709] at four. The entailment signals fall in the same direction, from
0.590 to 0.542 for the deployable variant and 0.645 to 0.592 for the oracle, but
their intervals overlap and we do not claim a decline for them. The reason they
cannot fall far is that they start close to chance: at two hops the deployable
variant is already at 0.590, so there is little room left. The clean version of
the dose-response test appears in Section~\ref{sec:decomp}, where removing the
mechanism lifts every hop count to the same value.

% ============================================================================
\section{Sufficiency, not length}
\label{sec:sufficiency}

\begin{figure}[t]
\centering
\begin{tikzpicture}
\begin{axis}[
  ybar, width=0.86\columnwidth, height=4.6cm, bar width=15pt,
  symbolic x coords={{gold+gold},{gold+distr.},{distr.+distr.}},
  xtick=data, ymin=0, ymax=0.72, ylabel={mean $P(\text{entail})$},
  axis y line*=left,
  tick label style={font=\scriptsize}, label style={font=\footnotesize},
  ymajorgrids, grid style={slate!15},
  nodes near coords, point meta=explicit symbolic,
  every node near coord/.append style={font=\tiny},
]
\addplot[fill=forest!45, draw=forest!80!black] coordinates {
  ({gold+gold},0.540) [0.540] ({gold+distr.},0.127) [0.127]
  ({distr.+distr.},0.025) [0.025] };
\end{axis}
\begin{axis}[
  width=0.86\columnwidth, height=4.6cm,
  symbolic x coords={{gold+gold},{gold+distr.},{distr.+distr.}},
  xtick=\empty, ymin=120, ymax=220, ylabel={premise length (words)},
  axis y line*=right, axis x line=none,
  tick label style={font=\scriptsize}, label style={font=\footnotesize},
  legend style={font=\scriptsize, at={(0.97,0.97)}, anchor=north east,
                draw=slate!40, fill=white},
]
\addplot[thick, coral, mark=*, mark size=2] coordinates {
  ({gold+gold},148.2) ({gold+distr.},172.1) ({distr.+distr.},197.6) };
\addlegendentry{premise length}
\end{axis}
\end{tikzpicture}
\caption{HotpotQA, 200 questions, all premises two paragraphs. Entailment falls
as premise length rises, so the set-level effect is not a length artefact. What
matters is whether both hops are present.}
\label{fig:length}
\end{figure}
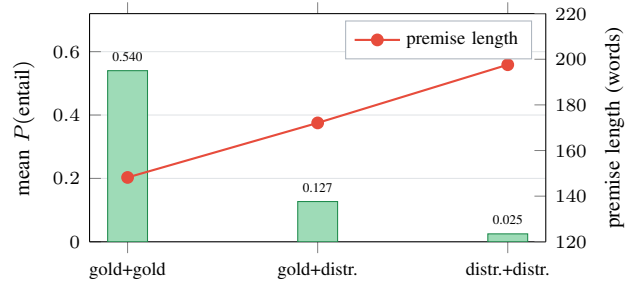

If per-chunk scoring fails because a chunk is not a sufficient premise, then
supplying both hops should repair it. It does. Holding the model, the questions
and the \textsc{oracle} hypothesis fixed and changing only the premise, scoring
single chunks gives 0.664 AUC and scoring the concatenation of both gold
paragraphs against other pairs gives \textbf{0.881}, a lift of $+0.217$. A gold
chunk alone receives 0.103 mean entailment probability; a gold pair receives
0.560.

The obvious alternative is that longer premises simply score higher.
Figure~\ref{fig:length} rules it out in the strongest available direction: the
relationship runs backwards. Gold pairs are the \emph{shortest} premises at 148
words and score 0.540. Gold with a distractor averages 172 words and scores
0.127. Two distractors average 198 words and score 0.025. More text, less
entailment.

The gold-plus-distractor row matters most. It contains a gold paragraph, it is
longer than the gold pair, and it still scores four times lower. Half the
evidence is not most of the way to the answer.

The deployable \textsc{slot} hypothesis behaves the same way at set level,
reaching 0.811 AUC over pairs drawn from the top five retrieved chunks against
0.620 per chunk. The gap to its 0.848 oracle is $0.037$, so the set-level effect
does not depend on seeing the answer.

% ============================================================================
\section{The single-hop control}
\label{sec:singlehop}

\begin{table}[t]
\centering
\caption{Single-hop control. SQuAD v1.1, 300 questions, nine same-article
distractors each, identical models and hypothesis construction.}
\label{tab:squad}
\begin{tabular}{@{}lcc@{}}
\toprule
\textbf{Signal} & \textbf{SQuAD} & \textbf{HotpotQA}\\
\midrule
$\semb$ embedding cosine           & 0.933 & 0.887\\
$\sent$ \textsc{slot} (deployable) & \textbf{0.927} & 0.643\\
$\sent$ \textsc{oracle}            & \textbf{0.951} & 0.669\\
\midrule
mean $P(\text{entail})$, gold       & 0.693 & 0.103\\
mean $P(\text{entail})$, distractor & 0.015 & n/a\\
\bottomrule
\end{tabular}
\end{table}

Everything so far is consistent with a competing account we have not yet
excluded: that entailment scoring of retrieved passages simply does not work at
this model scale, and multi-hop structure is incidental.

Table~\ref{tab:squad} excludes it. On SQuAD, where the gold paragraph contains
the answer and distractors come from the same article, the identical pipeline
reaches 0.951 AUC with \textsc{oracle} and 0.927 with the deployable
\textsc{slot} hypothesis. Gold paragraphs receive 0.693 mean entailment
probability, distractors 0.015.

Two things follow. Entailment verification works, and works well, when the
sufficiency assumption holds. And the deployable variant sits 0.024 behind the
oracle on SQuAD and 0.026 behind on HotpotQA, so the hypothesis template is
close to its ceiling in both cases; the ceilings differ by 0.282.

% ============================================================================
\section{Not the threshold, not the embedder}
\label{sec:threshold}

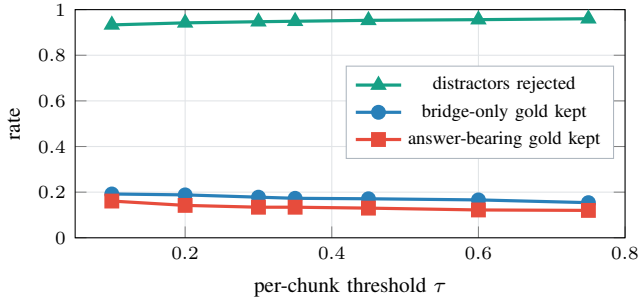
\begin{figure}[t]
\centering
\begin{tikzpicture}
\begin{axis}[
  width=\columnwidth, height=4.6cm,
  xlabel={per-chunk threshold $\tau$}, ylabel={rate},
  xmin=0.05, xmax=0.80, ymin=0, ymax=1.0,
  tick label style={font=\scriptsize}, label style={font=\footnotesize},
  grid=both, grid style={slate!15},
  legend style={font=\scriptsize, at={(0.98,0.55)}, anchor=east,
                draw=slate!40, fill=white},
]
\addplot[very thick, teal, mark=triangle*, mark size=2.4] coordinates {
  (0.10,0.933) (0.20,0.942) (0.30,0.947) (0.35,0.949) (0.45,0.953)
  (0.60,0.956) (0.75,0.960) };
\addlegendentry{distractors rejected}
\addplot[very thick, ocean, mark=*, mark size=2.2] coordinates {
  (0.10,0.192) (0.20,0.188) (0.30,0.178) (0.35,0.173) (0.45,0.171)
  (0.60,0.166) (0.75,0.154) };
\addlegendentry{bridge-only gold kept}
\addplot[very thick, coral, mark=square*, mark size=2.2] coordinates {
  (0.10,0.161) (0.20,0.142) (0.30,0.134) (0.35,0.134) (0.45,0.130)
  (0.60,0.122) (0.75,0.120) };
\addlegendentry{answer-bearing gold kept}
\end{axis}
\end{tikzpicture}
\caption{HotpotQA. Sweeping the per-chunk threshold moves nothing. Even at
$\tau=0.10$, 84\% of answer-bearing gold paragraphs are already rejected. The
entailment score distribution is collapsed near zero, so there is no operating
point to find.}
\label{fig:sweep}
\end{figure}

\paragraph{No threshold works}
A natural objection is that our per-chunk gate simply used a bad cutoff.
Figure~\ref{fig:sweep} sweeps it. Raising $\tau$ from 0.10 to 0.75 gains 2.7
points of distractor rejection and costs 4.0 points of gold recall, and the
curves are nearly flat throughout. More decisively, at the most permissive
threshold tested, only 16.1\% of answer-bearing gold paragraphs pass at all. The
score distribution is compressed against zero, so thresholding is close to
inoperative. There is no operating point being missed.

\paragraph{Not the embedder}
\begin{table}[t]
\centering
\caption{The directional deficit across three embedding models. HotpotQA,
200 questions.}
\label{tab:retrievers}
\begin{tabular}{@{}lccc@{}}
\toprule
\textbf{Embedder} & \makecell{\textbf{answer-}\\\textbf{bearing}} &
\makecell{\textbf{bridge}\\\textbf{only}} & \textbf{deficit}\\
\midrule
BAAI/bge-small-en-v1.5 & 0.821 & 0.937 & $+0.116$\\
intfloat/e5-base-v2    & 0.834 & 0.941 & $+0.107$\\
thenlper/gte-base      & 0.798 & 0.948 & $+0.150$\\
\bottomrule
\end{tabular}
\end{table}

Table~\ref{tab:retrievers} repeats the split with three widely used embedding
models spanning two architectures and three training recipes. Every one prefers
the bridge paragraph by 0.107 to 0.150. The directional bias is a property of
conditioning on the question, not of any particular encoder.

% ============================================================================
\section{The structural signal}
\label{sec:hrr}

The HRR signal fails for the same reason as entailment and for a second reason
on top of it, and we treat it briefly because it is not central. It has the
largest directional deficit of any signal, $+0.136$, so it inherits the same
blindness. Separately, it faces an extraction dilemma. Our extractor takes the
first filler for each of four roles, pinning $k$, the bindings per trace, at 3.98
against 41.9 candidate pairs available per paragraph, discarding 90.5\% of the
structure. Binding all of it does not help: for a trace of $k$ bindings with
near-orthogonal roles, unbinding returns the target filler plus $k-1$ crosstalk
terms of comparable norm, so
$\mathbb{E}[\cossim(\mathbf{v}_r^{\dagger}\conv T(c),\mathbf{v}_{\phi_c(r)})]
\approx k^{-1/2}$, which is 0.154 at $k=42$ and below any usable threshold.
Paragraph-level HRR is squeezed from both sides. An earlier version of this work
blamed a capacity threshold in the generator; that was wrong, since the
generator never sees the structural score in a form it could exploit.

% ============================================================================
\section{What it costs, and what conditioning recovers}
\label{sec:endtoend}

\begin{algorithm}[t]
\caption{Conditional evidence selection}
\label{alg:cond}
\begin{algorithmic}[1]
\Require question $q$, candidates $\mathcal{C}$, budget $k$
\State $\mathcal{T}\gets$ top-$k$ of $\mathcal{C}$ by $\semb(q,\cdot)$
\State $c_1\gets\arg\max_{c\in\mathcal{T}}\semb(q,c)$
  \Comment{hop 1; gold 92\% of the time}
\State $h\gets\mathcal{H}(q)$ \Comment{object-level proposition, no gold answer}
\ForAll{$c_j\in\mathcal{T}\setminus\{c_1\}$}
  \State $u_j\gets P\big(\text{entail}\mid c_1\oplus c_j,\ h\big)$
    \Comment{$c_1$ makes the premise sufficient}
\EndFor
\State $c_2\gets\arg\max_j u_j$
\State \Return $\{c_1,c_2\}$
  \Comment{$k-1$ calls, against $\binom{k}{2}$ for blind pair search}
\end{algorithmic}
\end{algorithm}

\begin{table}[t]
\centering
\caption{Paired exact McNemar against B1, unfiltered retrieval. $b$ and $c$ are
discordant counts; $p$ is Holm-adjusted within each dataset. Qwen2.5-1.5B,
standard prompt. $^{*}p<.05$, $^{**}p<.01$, $^{***}p<.001$.}
\label{tab:tests}
\setlength{\tabcolsep}{4pt}
\begin{tabular}{@{}llrrrr@{}}
\toprule
\textbf{Dataset} & \textbf{Selector} & $\Delta$\textbf{EM} & $b$ & $c$ & \textbf{Holm }$p$\\
\midrule
HotpotQA & B5   & $+8.0$  & 68 & 28 & $<.001^{***}$\\
         & B4   & $+0.8$  & 35 & 31 & 1.000\\
         & COND & $-1.2$  & 37 & 43 & 1.000\\
         & SET  & $-2.6$  & 32 & 45 & 0.513\\
         & B2   & $-6.2$  & 20 & 51 & $0.001^{**}$\\
         & PC   & $-13.4$ & 31 & 98 & $<.001^{***}$\\
\midrule
2Wiki    & B5   & $+8.2$  & 76 & 35 & $0.001^{***}$\\
         & B2   & $+0.6$  & 49 & 46 & 1.000\\
         & B4   & $+0.2$  & 48 & 47 & 1.000\\
         & COND & $-1.4$  & 43 & 50 & 1.000\\
         & SET  & $-1.6$  & 47 & 55 & 1.000\\
         & PC   & $-6.2$  & 42 & 73 & $0.025^{*}$\\
\midrule
MuSiQue  & B5   & $+20.1$ & 61 & \ 9 & $<.001^{***}$\\
         & B4   & $-3.1$  & 15 & 23 & 0.308\\
         & B2   & $-3.9$  & 15 & 25 & 0.308\\
         & COND & $-6.6$  & 10 & 27 & $0.023^{*}$\\
         & SET  & $-7.3$  & \ 9 & 28 & $0.010^{*}$\\
         & PC   & $-11.2$ & \ 9 & 38 & $<.001^{***}$\\
\bottomrule
\end{tabular}
\end{table}

\begin{table}[t]
\centering
\caption{End to end, Qwen2.5-1.5B, greedy decoding. \emph{rec} is the fraction
of available gold paragraphs reaching the generator. Brackets are 95\% Wilson
intervals. MuSiQue is restricted to its 2-hop questions, since pair selectors
cannot reach a 3- or 4-hop answer by construction.}
\label{tab:endtoend}
\setlength{\tabcolsep}{3pt}
\begin{tabular}{@{}llccc@{}}
\toprule
\textbf{Data} & \textbf{Selector} & \textbf{EM} & \textbf{95\% CI} & \textbf{rec}\\
\midrule
\multirow{7}{*}{\makecell[l]{HotpotQA\\$n=500$}}
 & B5 oracle gold pair    & \textbf{55.0} & [50.6,\,59.3] & 1.00\\
 & B4 reranker top-2      & 47.8 & [43.5,\,52.2] & 0.86\\
 & B1 all five, no filter & 47.0 & [42.7,\,51.4] & 0.92\\
 & \textbf{COND} (ours)   & 45.8 & [41.5,\,50.2] & 0.76\\
 & SET blind pair search  & 44.4 & [40.1,\,48.8] & 0.68\\
 & B2 top-2 by retriever  & 40.8 & [36.6,\,45.2] & 0.77\\
 & PC per-chunk gate      & 33.6 & [29.6,\,37.9] & 0.46\\
\midrule
\multirow{7}{*}{\makecell[l]{2Wiki\\$n=500$}}
 & B5 oracle gold pair    & \textbf{42.2} & [38.0,\,46.6] & 1.00\\
 & B2 top-2 by retriever  & 34.6 & [30.6,\,38.9] & 0.70\\
 & B4 reranker top-2      & 34.2 & [30.2,\,38.5] & 0.74\\
 & B1 all five, no filter & 34.0 & [30.0,\,38.3] & 0.87\\
 & \textbf{COND} (ours)   & 32.6 & [28.6,\,36.8] & 0.61\\
 & SET blind pair search  & 32.4 & [28.5,\,36.6] & 0.46\\
 & PC per-chunk gate      & 27.8 & [24.1,\,31.9] & 0.42\\
\midrule
\multirow{7}{*}{\makecell[l]{MuSiQue\\$n=259$}}
 & B5 oracle gold pair    & \textbf{40.9} & [35.1,\,47.0] & 1.00\\
 & B1 all five, no filter & 20.9 & [16.4,\,26.2] & 0.74\\
 & B4 reranker top-2      & 17.8 & [13.6,\,22.9] & 0.60\\
 & B2 top-2 by retriever  & 17.0 & [12.9,\,22.0] & 0.54\\
 & \textbf{COND} (ours)   & 14.3 & [10.6,\,19.1] & 0.51\\
 & SET blind pair search  & 13.5 & [\ 9.9,\,18.2] & 0.41\\
 & PC per-chunk gate      & \ 9.7 & [\ 6.6,\,13.9] & 0.35\\
\bottomrule
\end{tabular}
\end{table}

An AUC gap is not an answer-quality gap, so we measured both. Seven selectors
feed the same generator: no filtering, truncation to top two by retriever score,
a cross-encoder reranker~\cite{xiao2024bge}, the per-chunk gate this paper has
been analysing, blind set-level search over all $\binom{5}{2}$ pairs, the
conditional selector of Algorithm~\ref{alg:cond}, and an oracle supplying the
gold paragraphs.

Table~\ref{tab:tests} gives paired exact McNemar tests against B1, unfiltered
retrieval, Holm-corrected within each dataset. B1 is the right reference because
the question is whether filtering helps at all.

\paragraph{Per-chunk gating is significantly worse than not filtering}
On all three datasets: $-13.4$ Exact Match on HotpotQA ($p<.001$), $-6.2$ on
2Wiki ($p=.025$), $-11.2$ on MuSiQue ($p<.001$). The mechanism is in the recall
column: the gate keeps 1.2 chunks on average and lets only 35\% to 46\% of gold
paragraphs through. This is the strongest form our claim takes.

Conditioning recovers evidence at lower cost, though not accuracy.
Algorithm~\ref{alg:cond} beats blind pair search on gold recall by $+0.072$,
$+0.149$ and $+0.078$, using $k-1=4$ entailment calls against $\binom{5}{2}=10$,
a 60\% reduction. Its anchor lands on a gold paragraph 92.0\% of the time on
HotpotQA and 96.8\% on 2Wiki, but only 77.8\% on MuSiQue, whose twenty
candidates make the first hop harder to fix; that is why it helps least there.

Against B1 the accuracy picture is worse than we first read it. On HotpotQA and
2Wiki the conditional selector is indistinguishable from no filtering ($-1.2$,
$p=1.0$; $-1.4$, $p=1.0$). On MuSiQue it is \emph{significantly worse}
($-6.6$, $p=.023$), as is blind pair search ($-7.3$, $p=.010$). Our own proposal
does not survive its own test on the hardest dataset, and we report that rather
than resting on the recall gain.

No selector is significantly better than B1 anywhere. The reranker comes closest
and never separates ($+0.8$, $+0.2$, $-3.1$; all $p>.3$). Truncation to two
chunks is significantly worse on HotpotQA ($-6.2$, $p=.001$). Only the oracle
beats B1, by $+8.0$, $+8.2$ and $+20.1$, all $p<.001$.

The headroom is real and unclaimed. The oracle sits 7.2 points above the best
deployable selector on HotpotQA, 7.6 on 2Wiki and \textbf{20.1} on MuSiQue,
where it nearly doubles the best deployable score. None of the verifiers we
tested collect any of it.

\paragraph{Why good AUC did not produce good answers}
Set-level pair AUC is 0.811, well above 0.620 per chunk. But a deployed selector
is an argmax, not a ranking, and the argmax lands on the correct gold pair only
44.4\% of the time, though it contains at least one gold paragraph 92.2\% of the
time. That is why set-level scoring beats the per-chunk gate and still trails no
filtering. AUC is necessary for a verifier and it is not sufficient; reporting it
without argmax accuracy overstates what a signal will do in place.

% ============================================================================
\section{What repairs it}
\label{sec:decomp}

\begin{figure}[t]
\centering
\begin{tikzpicture}
\begin{axis}[
  width=\columnwidth, height=5.0cm,
  xlabel={hops required to answer}, ylabel={AUC on the supporting paragraph},
  xmin=1.7, xmax=4.3, ymin=0.45, ymax=0.95, xtick={2,3,4},
  tick label style={font=\scriptsize}, label style={font=\footnotesize},
  grid=both, grid style={slate!15},
  legend style={font=\scriptsize, at={(0.98,0.5)}, anchor=east,
                draw=slate!40, fill=white},
  nodes near coords, point meta=explicit symbolic,
  every node near coord/.append style={font=\tiny, yshift=2pt},
]
\addplot[very thick, forest, mark=square*, mark size=2.4] coordinates {
  (2,0.848) [.848] (3,0.849) [.849] (4,0.804) [.804] };
\addlegendentry{gold sub-question}
\addplot[very thick, amber, mark=triangle*, mark size=2.8] coordinates {
  (2,0.710) [.710] (3,0.556) [.556] (4,0.560) [.560] };
\addlegendentry{Qwen2.5-7B, anchored}
\addplot[very thick, coral, mark=*, mark size=2.4] coordinates {
  (2,0.570) [.570] (3,0.503) [.503] (4,0.558) [.558] };
\addlegendentry{original question}
\addplot[dashed, slate, forget plot] coordinates {(1.7,0.5) (4.3,0.5)};
\node[font=\scriptsize, text=slate, anchor=west] at (axis cs:1.75,0.472) {chance};
\end{axis}
\end{tikzpicture}
\caption{MuSiQue, 9997 question-paragraph pairs. Asking whether a paragraph
supports a later hop is near chance when the hypothesis comes from the original
question and becomes tractable when it comes from the sub-question. The gold
curve is flat within its intervals. An off-the-shelf decomposer matches most of
the gain at two hops and falls back to the original-question level beyond that;
the drop from 0.710 [0.676, 0.743] to 0.556 [0.513, 0.594] lies outside the
intervals.}
\label{fig:decomp}
\end{figure}
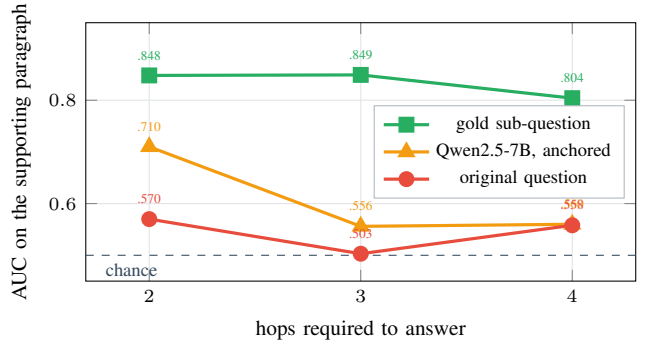

Everything so far says per-chunk verification fails because the hypothesis is
built from a question whose answer no single chunk supports. That predicts a
repair: build the hypothesis from the sub-question instead.

MuSiQue lets us test this without building a decomposer. It ships
\texttt{question\_decomposition}, giving each hop's sub-question and the index of
the paragraph that supports it. We take a later hop, resolve its placeholders
against earlier answers, and ask a single question of the entailment model:
which paragraph supports \emph{this} hop? We compare three hypotheses, all
scored against the same 9997 paragraph candidates.

\begin{table}[t]
\centering
\caption{Identifying the paragraph that supports a later hop. MuSiQue, 500
questions, 9997 candidate paragraphs, identical NLI model throughout. The
decomposer is Qwen2.5-7B-Instruct with no fine-tuning and no gold annotation at
inference. Percentages are of the distance from the original question to the
gold sub-question.}
\label{tab:decomp}
\begin{tabular}{@{}lc@{}}
\toprule
\textbf{Hypothesis built from} & \textbf{AUC \small{[95\% CI]}}\\
\midrule
the original question                 & 0.546 \small{[0.523, 0.569]}\\
the decomposed sub-question           & \textbf{0.840} \small{[0.824, 0.856]}\\
the sub-question with its gold answer & 0.936 \small{[0.924, 0.947]}\\
\midrule
\multicolumn{2}{@{}l}{paired lift: $\mathbf{+0.355}$ \small{[0.331, 0.382]}}\\
\bottomrule
\end{tabular}
\end{table}

Table~\ref{tab:decomp} and Figure~\ref{fig:decomp} give the result.

The original question gives 0.546 AUC for identifying the paragraph a later hop
depends on, which is chance. That is the sharpest form of this paper's negative
claim: a verifier conditioned on the question as asked cannot tell which
paragraph carries the second hop.

The gold sub-question gives 0.840, a paired lift of $+0.355$ with an interval of
$[0.331, 0.382]$. The same NLI model, the same paragraphs and the same task
become tractable purely by changing what the hypothesis asks about.

Hop count stops mattering under decomposition. With the original question the
AUC is 0.570, 0.503 and 0.558 at two, three and four hops, with overlapping
intervals and no usable value at any depth. With the gold sub-question it is
0.848 [0.826, 0.869], 0.849 [0.823, 0.873] and 0.804 [0.750, 0.850]: flat within
its intervals, and every interval excluding the whole original-question range. A
Depth stops predicting difficulty once the verifier is pointed at the hop in
question.

The lift is a paired bootstrap on the difference, so it is not an artefact of
comparing two separately estimated quantities.

\paragraph{An off-the-shelf decomposer captures a third of it}
The gold annotation makes this a ceiling. To see how much of it is reachable, we
had Qwen2.5-7B-Instruct produce the decomposition with no fine-tuning and no
gold annotation, in two settings. Given only the question, it reaches 0.533
[0.510, 0.557], which is \emph{below} the original-question baseline. Given the
question and the top retrieved paragraph, in the shape Self-Ask uses, it reaches
0.637 [0.611, 0.662], a paired lift of $+0.091$ [0.066, 0.115] and 31\% of the
ceiling.

The blind setting hurts and the anchored setting helps, which fits
the rest of the results. A decomposer conditioned only on the original question
has the same blind spot the verifier does; it needs retrieved text to work from.

Two things bound the deployable version. Its gain is gated by whether the
retrieval anchor is correct: 0.661 on the 389 questions where the top chunk is
gold and 0.546 on the 111 where it is not, the latter being exactly the
original-question baseline. And it works at two hops, 0.710 [0.676, 0.743], then
falls to 0.556 [0.513, 0.594] and 0.560 [0.499, 0.614] at three and four, a drop
outside the intervals. Our prompt asks for one follow-up question, which is well
posed when a single hop remains and ambiguous when three do. Iterative
decomposition, one hop at a time, would address that. We have not run it.

The effect comes from restructuring the question and not from having more text
available. We also tested the naive intervention of appending the top retrieved
chunk to the query at three granularities, title, first sentence and full
paragraph, and re-scoring. Mean lift on answer-bearing gold was $+0.034$ across
datasets, and on HotpotQA the full-paragraph variant \emph{hurt} by $0.028$. A
deployable approximation that conditions the hypothesis on the anchor title
recovered $+0.059$, $+0.042$ and $-0.034$ on the three datasets, a small
fraction of the oracle's $+0.355$.

Producing decompositions is not an open problem: Self-Ask and IRCoT generate them
already, for retrieval. What we report is that the same artefact repairs
verification, that nobody appears to be using it there, and that a
general-purpose model reaches a third of the available gain without being asked
to do anything special. The remaining 0.20 AUC is what a decomposer built for
this would be worth.

\footnote{A measurement artefact in our own tooling is worth recording, since it
is the effect this paper documents. An earlier version of this experiment forced
every gold sub-question into a relational template, producing hypotheses of the
form ``The answer of \emph{[question]} is something.'' Most MuSiQue
sub-questions are natural language rather than relational, so that template
turned them into meta-statements about an answer, and the measured AUC fell from
0.840 to 0.779. That is the out-of-distribution effect described in
Section~\ref{sec:setup}, showing up in our own measurement code.}

% ============================================================================
\section{Robustness}
\label{sec:robust}

\begin{table}[t]
\centering
\caption{The end-to-end ordering across three generator sizes. Rank 1 is best.
Per-chunk gating is last in every cell but one, and its penalty grows with
capability.}
\label{tab:scale}
\setlength{\tabcolsep}{4pt}
\begin{tabular}{@{}lccc@{}}
\toprule
\textbf{Dataset} & \textbf{0.5B} & \textbf{1.5B} & \textbf{3B}\\
\midrule
\multicolumn{4}{@{}l}{\emph{PC minus best deployable selector, Exact Match}}\\
HotpotQA & $-4.6$ & $-14.2$ & $\mathbf{-19.4}$\\
2Wiki    & $\ \ 0.0$ & $-6.8$ & $\mathbf{-13.6}$\\
MuSiQue  & $-1.9$ & $-11.2$ & $-8.1$\\
\midrule
\multicolumn{4}{@{}l}{\emph{oracle minus best deployable selector, Exact Match}}\\
HotpotQA & $-0.6$ & $+7.2$ & $\mathbf{+10.6}$\\
2Wiki    & $-1.0$ & $+7.6$ & $+7.2$\\
MuSiQue  & $+5.0$ & $+20.1$ & $\mathbf{+23.6}$\\
\bottomrule
\end{tabular}
\end{table}

\paragraph{Generator scale}
Table~\ref{tab:scale} repeats the seven-selector grid at 0.5B, 1.5B and 3B. Two
patterns hold across all three datasets. First, the cost of per-chunk gating
\emph{grows} with generator capability, from 4.6 Exact Match at 0.5B to 19.4 at
3B on HotpotQA. A stronger model is hurt more by deleted evidence, because it
could have used it. Second, the oracle gap grows the same way, reaching 23.6
points on MuSiQue at 3B. Better selection becomes more valuable, not less, as
generators improve.

At 0.5B nothing is distinguishable. All seven selectors land within a few points
with fully overlapping intervals, and on two datasets the oracle is not even
best. A 0.5B generator handed both gold paragraphs answers 10.8\% of HotpotQA
and 7.7\% of MuSiQue. Selection cannot matter when the generator cannot use the
selection, and any verification result reported at that scale on these tasks is
measuring noise.

\paragraph{Prompt}
Repeating the comparison with a differently framed generation prompt moves every
absolute number down by 8 to 10 Exact Match and leaves the ordering intact:
oracle, then reranker, then the conditional selector and unfiltered retrieval
within noise of each other, then per-chunk gating last, on all three datasets.

\paragraph{Answer-matching criterion}
The answer-bearing split underpins Sections~\ref{sec:direction}
and~\ref{sec:sufficiency}, so we recomputed it three ways: exact normalised
string match, match against MuSiQue's answer aliases, and token overlap at 0.8.
The directional deficit moves by at most 0.007 on any dataset. The split is not
an artefact of how answer-bearing is operationalised.

% ============================================================================
\section{Discussion}
\label{sec:discussion}

The recommendation we hold with most confidence is negative: do not gate
multi-hop retrieval per chunk. Entailment filtering is the worst of seven
selectors on three datasets and it fails by removing the evidence the question
needs. A system that adds such a gate will appear to be doing careful work while
deleting the second hop.

If some selection is required, anchor before searching. The embedding signal is
reliable on the first hop, at 0.92 to 0.97 AUC on question-named evidence, and
unreliable on the second, so Algorithm~\ref{alg:cond} uses each signal where it
works: embedding to fix the anchor, entailment to score completions given that
anchor. It recovers 7 to 15 points of gold recall over blind search at 40\% of
the verification cost. We report the design because the evidence points at it,
not because this instance of it beats doing nothing. It does not.

Self-Ask~\cite{press2023selfask} and IRCoT~\cite{trivedi2023ircot} decompose the
query between hops because the second hop is not reachable from the original
question. They use the decomposition for retrieval and discard it before
verification. Section~\ref{sec:decomp} indicates that discarding it is what
breaks the verifier: the same artefact lifts entailment on a later hop from
chance to 0.840 and flattens the hop-count effect.

The integration is cheap. A pipeline already running Self-Ask or IRCoT has the
sub-questions in hand, and passing them to the verifier instead of the original
query costs one string substitution. The ceiling is $+0.355$ AUC and an
off-the-shelf 7B decomposer reaches $+0.091$ of it without any adaptation, so
about two thirds of the distance is still open. Two specific things would close
part of it: decomposing iteratively rather than in one shot, since the gain
survives only at two hops, and improving first-hop retrieval, since the gain
vanishes entirely when the anchor is wrong.

On method: the AUC measurements cost about three dollars of cloud compute and
showed the per-chunk signal was unusable before a single answer was generated.
Separation alone would have misled us in the other direction, though. A pair AUC
of 0.811 looked like a fix and became a selector that still loses to no
filtering, because a deployed selector is an argmax and the argmax is right
44.4\% of the time. Both measurements are cheap and neither substitutes for the
other.

Both the cost of bad selection and the value of good selection grow with
generator scale (Table~\ref{tab:scale}). Per-chunk gating costs 4.6 Exact Match
at 0.5B and 19.4 at 3B, and the oracle gap widens from $-0.6$ to $+10.6$ on
HotpotQA, reaching 23.6 on MuSiQue. Evidence selection for multi-hop RAG becomes
more valuable as generators improve.

We would encourage anyone running a comparison like this to include the oracle
row. Without it we would have read the reranker's 47.8 as near-ceiling and the
separation gain as the main thing left to chase. It costs one extra
configuration and it repriced every other row in the table.

\paragraph{Single-hop deployments are a different case}
The deployable \textsc{slot} variant reaches 0.927 AUC on SQuAD, close to its
0.951 oracle, at 184\,M parameters. Where one passage carries the whole answer,
entailment gating rests on an assumption that holds. Everything negative here is
specific to multi-hop.

% ============================================================================
\section{Limitations}
\label{sec:limits}

The decomposition ceiling uses MuSiQue's gold annotations, and while we measure
what an off-the-shelf decomposer reaches, we did not build one, did not run the
decomposed verifier end to end, and did not test transfer to HotpotQA or 2Wiki,
which ship no decompositions. The deployable measurement also inherits the
retrieval anchor: on the 111 questions where the top chunk is not gold it
provides nothing. The gaps among the top deployable selectors are inside their
intervals, so we order them only weakly.
The MuSiQue end-to-end rows cover its 2-hop subset, because pair selectors
cannot reach a 3- or 4-hop answer; the AUC analysis uses every hop count, and
selection over larger sets remains untested. Our conditional
selector is one design among many: beam search over larger sets, or rewriting
the hypothesis using the anchor rather than concatenating it, might do better
and we have not tried either. The single-hop control uses SQuAD, where the gold
paragraph is guaranteed to contain the answer, which is easier than open-domain
single-hop retrieval; the true single-hop number likely lies between 0.664 and
0.951. The role extractor is one implementation, though
the $k^{-1/2}$ crosstalk bound in Section~\ref{sec:hrr} limits how much a better
one could help. All entailment
numbers come from one model family.

% ============================================================================
\section{Conclusion}
\label{sec:conclusion}

We built a three-signal chunk verifier for retrieval-augmented generation and
measured, before evaluating it end to end, how well each signal separates gold
evidence from distractors. On three multi-hop datasets none of them worked:
embedding similarity reached 0.76 to 0.89, entailment never exceeded 0.67 even
when handed the gold answer, and structural matching reached 0.62.

The cause is an assumption rather than a defect. Per-chunk verification treats
each chunk as a sufficient premise, and in multi-hop retrieval that is false for
the chunk carrying the second hop. Seven measurements pin it down: the signals
prefer the paragraph named in the question, the preference vanishes on
comparison questions, it deepens monotonically with hop count, supplying both
hops lifts entailment from 0.664 to 0.881, premise length runs the opposite way,
single-hop SQuAD reaches 0.951, and no threshold or embedder changes any of it.

End to end this costs what the analysis predicts, across three datasets, three
generator sizes and two prompts. Per-chunk gating is worst in every cell, and its
penalty grows with generator capability, from 4.6 Exact Match at 0.5B to 19.4 at
3B. An oracle selector sits up to 23.6 points above the best deployable method,
and that gap widens with scale too.

What repairs it is changing the question the verifier is asked. Conditioning the
hypothesis on the decomposed sub-question rather than the original query lifts
entailment on a later hop from 0.546, which is chance, to 0.840, a paired lift of
$+0.355$ [0.331, 0.382]. Hop count then stops mattering: 0.848, 0.849 and 0.804
at two, three and four hops. An off-the-shelf 7B decomposer with no adaptation
reaches 0.637, a third of that ceiling, and only at two hops. Appending retrieved
text to the query does not work at all, so the effect comes from restructuring
the question rather than from added context. Iterative retrieval systems already
build these decompositions and discard them before verifying.

Verification is not broken. It is being asked to judge sufficiency one chunk at
a time, against a question that no single chunk can answer.

% ============================================================================
\section*{Reproducibility}

Code, the exact question identifiers for every split, per-question signal
traces, and the scripts that generate every table and figure here are at
\url{https://github.com/iamhero2709/verification-without-sufficiency}. All runs use greedy
decoding and seed 1337. Every experiment in this paper cost under fifteen
dollars of cloud compute and about five hours of wall clock.
Confidence intervals are question-stratified bootstraps with 1000 resamples and
the tables reporting them are emitted by script, not typed. Every number in
every table is produced by a script reading the released traces.

\balance

\end{document}